\documentclass{article}

\usepackage[final]{corl_2018} 
\usepackage{amsmath}
\usepackage{bm}
\usepackage{graphicx}
\usepackage{siunitx}
\usepackage{caption}
\usepackage{subcaption} 
\usepackage{wrapfig}
\newcommand{\tg}{\textsc{TG}}
\newcommand{\pmtg}{\textsc{PMTG}}
\graphicspath{ {figures/} }
\usepackage[dvipsnames]{xcolor}

\title{Policies Modulating Trajectory Generators}

\author{Atil Iscen\textsuperscript{1} \qquad Ken Caluwaerts\textsuperscript{1} \qquad Jie Tan\textsuperscript{2}  \qquad Tingnan Zhang\textsuperscript{2} \AND Erwin Coumans\textsuperscript{2} \qquad Vikas Sindhwani\textsuperscript{1} \qquad Vincent Vanhoucke\textsuperscript{2}\\
\\
\textsuperscript{1}Google Brain, New York, United States\\
\textsuperscript{2}Google Brain, Mountain View, United States\\
\texttt{\{atil, tensegrity, jietan, tingnan, erwincoumans, sindhwani, vanhoucke\}}\\
\texttt{@google.com}}

\begin{document}
\maketitle


\begin{abstract}
We propose an architecture for learning complex controllable behaviors by having simple Policies Modulate Trajectory Generators (\pmtg{}), a powerful combination that can provide both memory and prior knowledge to the controller. 
The result is a flexible architecture that is applicable to a class of problems with periodic motion for which one has an insight into the class of trajectories that might lead to a desired behavior. We illustrate the basics of our architecture using a synthetic control problem, then go on to learn speed-controlled locomotion for a quadrupedal robot by using Deep Reinforcement Learning and Evolutionary Strategies. 
We demonstrate that a simple linear policy, when paired with a parametric Trajectory Generator for quadrupedal gaits, can induce walking behaviors with controllable speed from $4$-dimensional IMU observations alone, and can be learned in under 1000 rollouts. 
We also transfer these policies to a real robot and show locomotion with controllable forward velocity.

\end{abstract}

\keywords{Reinforcement Learning, Control, Locomotion} 


\section{Introduction}
The recent success of Deep Learning (DL) on simulated robotic tasks has opened an exciting research direction. Nevertheless, many robotic tasks such as locomotion still remain an open problem for learning-based methods due to their complexity or dynamics. From a Deep Learning (DL) perspective, one way to tackle these complex problems is by using more and more complex policies (such as recurrent networks). Unfortunately, more complex policies are harder to train and require even more training data which is often problematic for robotics.

Robotics is naturally a great playground for combining strong prior knowledge with DL. The robotics literature contains many forms of prior knowledge about locomotion tasks and nature provides impressive examples of similar architectures. Note that this knowledge does not need to be in the form of perfect examples, it can also be in form of intuition about the specific robotic problem. As an example, for locomotion it can be defined as leg movements patterns based on certain gait and external parameters. 

\begin{figure}[t]
      \centering
      \includegraphics[width=0.6\columnwidth]{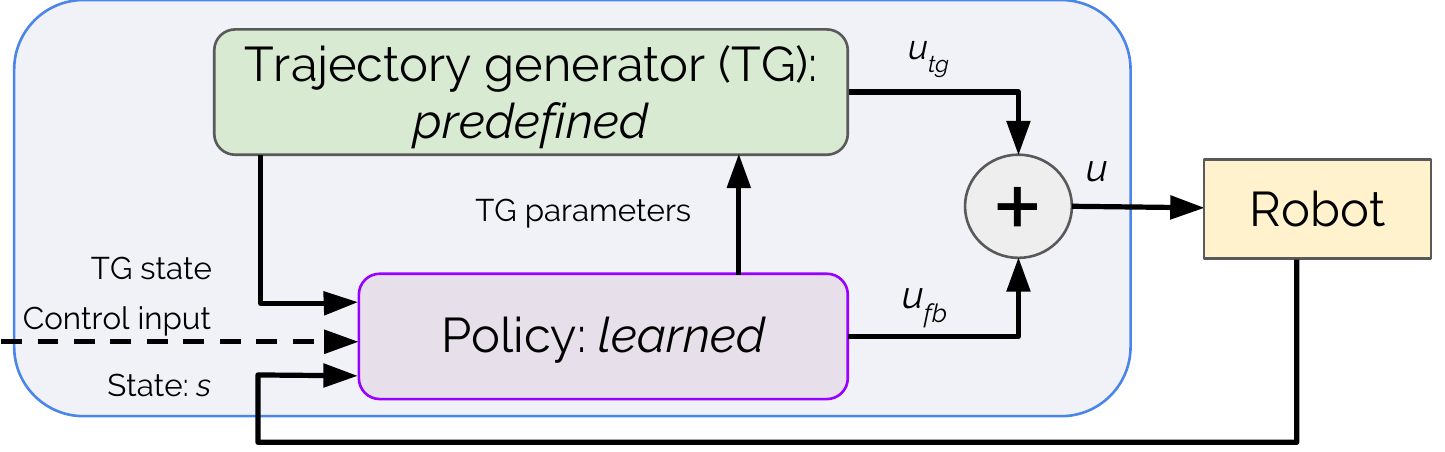}
      \caption{Overview of \pmtg{}: The output (actions) $u_{tg}$ of of a predefined Trajectory Generator (\tg{}) is combined with that of a learned policy network ($u_{fb}$). The learned policy also modulates the parameters of the \tg{} at each time step and observes its state.}
      \label{fig:architecture}
\end{figure}

We incorporate this intuitive type of prior knowledge into learning in the form of a parameterized Trajectory Generator (\tg{}). We keep the \tg{} separate from the learned policy and define it as a stateful module that outputs actions $\boldsymbol{u}_{tg}$ (e.g. target motor positions) which depend on its internal state and external parameters. We introduce a new architecture in which the policy has control over the \tg{} by modulating its parameters as well as directly correcting its output (Fig.~\ref{fig:architecture}). In exchange, the policy receives the \tg{}'s state as part of its observation. 
As the \tg{} is stateful, these connections yield a controller that is implicitly recurrent while using a feed-forward Neural Network (NN) as the learned policy. The advantage of using a feed-forward NN is that learning is often significantly less demanding than with recurrent NNs. Moreover, this separation of the feed-forward policy and the stateful \tg{} makes the architecture compatible with any reward based learning method. 

We call our architecture \emph{Policies Modulating Trajectory Generators} (\pmtg{}) to stress the interaction between the learned policy and the predefined \tg{}. In essence, we replace the task of learning a locomotion controller by that of learning to modulate a \tg{} in parallel with learning to control a robot. 

In this manuscript, we first illustrate the architecture of \pmtg{} using a synthetic control problem. Next, we tackle quadruped locomotion using \pmtg{}. We use desired speed as the control input and different \tg{}s that generate leg trajectories based on parameters such as stride length, frequency and walking height. We train our policies in simulation using Reinforcement Learning (RL) or Evolutionary Strategies (ES) with policies as simple as one linear layer using only a four-dimensional proprioceptive observation space (IMU). Finally we transfer the learned policies to a real robot and demonstrate learned locomotion with controllable speed. 

\section{Related Work}\label{sec:related}
Optimization is an effective tool to automatically design locomotion controllers. Popular techniques include Black-box Optimization~\cite{icra18-minitaur}, Bayesian Optimization~\cite{calandra2016bayesian}, Evolutionary Algorithms~\cite{such2017deep, mania2018simple} and Reinforcement Learning~\cite{HeessTSLMWTEWER17, DBLP:journals/corr/SchulmanWDRK17, tan2018}. In recent works, neural networks are commonly used parameterizations of the control policy. While the architecture of the neural network plays an important role in learning, simple fully-connected feed-forward networks are often used due to the challenges to design network architecture. Although several prior works~\cite{stanley2002evolving, real2017large, zoph2016neural} can automatically search for the optimal architecture, they require tremendous amounts of time and computation. In computer graphics, special architectures, such as Phase-Functioned Neural Networks~\cite{holden2017phase} and Mode-Adaptive Neural Networks~\cite{Zhang2018}, have been proposed to synthesize locomotion controllers from motion capture data. While these methods generate vivid animations, they do not consider physics and balance control, and thus, are not suitable for robotics.

Locomotion is periodic and structured motion. Classical locomotion controllers often use a cyclic open loop trajectory, such as sine curves~\cite{tan2018,iscen2013controlling} or Central Pattern Generators (CPG)~\cite{ijspeert2008central} to parameterize the movement of each actuated degrees of freedom. To control locomotion, Gay et al.~\cite{gay2013learning} learned a neural network that modulated a CPG. Sharma and Kitani~\cite{Sharma-2018-103569} designed phase-parametric policies by exploiting the cyclic nature of locomotion. Tan et al. \cite{tan2018} learnt a feedback balance controller on top of a user-specified pattern generator. Although our method is inspired by \cite{tan2018}, there are key differences. In Tan et al. \cite{tan2018}, the pattern generator is fixed and independent to the feedback control. In this way, the feedback control can only modify the gait in the vicinity of the signal defined by pattern generator. In contrast, in our architecture, the feedback control modulates the \tg{} including its frequency. This is crucial since we are interested in dynamically changing the high-level behavior of the locomotion, which requires changing the underlying trajectory and its frequency.

In this paper, our focus is to learn controllable locomotion, in which the robot can change its behavior given external control signals (e.g. changing running speed with a remote). One way to achieve it is to train separate controllers for corresponding behaviors and switch them online according to the user-specified signals~\cite{peng2018deepmimic}. However, abruptly switching controllers can cause jerky motion and loss of balance. An alternative is to learn a generic controller and add the external control signals to the observation space~\cite{tan2014learning}. As only one controller is learned and there is no need for transitions, this formulation significantly decreases the difficulty of the task. We choose the second approach and show that with \pmtg{}, we can learn controllable locomotion efficiently.


\section{Architecture}
\label{sec:method}
Our basic architecture is shown in Fig.~\ref{fig:architecture} and consists of three main blocks: an existing/predefined controller, a learned policy, and the system to control (a robot).
In this manuscript, we refer to the existing controller as the Trajectory Generator (\tg{}), because we limit our experimental section to periodic motions. However, \pmtg{} can be extended to various types of predefined controllers (e.g. a kinematic controller) in a straightforward manner.

Just like the robotic system to control, the \tg{} is a black box from the policy's point of view. It receives a number of parameters as inputs from the policy and it outputs its state and actions at every time step. Hence, learning a policy in \pmtg{} is equivalent to learning to control the original dynamical system (robot) extended by the \tg{}. One simply concatenates the action space of the original problem and the controllable parameters of the \tg{}. Similarly, the observation space is extended with the state variables of the \tg{}. Note, that the state of the \tg{} does not affect the reward. 

The policy can optionally accept control inputs to allow external control of the robot. These control inputs are also appended to the robot's observations/state and fed into the policy. This simple formulation allows \pmtg{} to be trained using a large selection of policy optimization methods.

The outputs of the controller are the actions $\boldsymbol{u}$ that control the robot's actuators. These actions are computed as the sum of the output of the \tg{} and the policy\footnote{We use the subscript ${~}_{fb}$ to refer to the policy because it computes feedback signals.}: 
\begin{equation*}
    \boldsymbol{u}=\boldsymbol{u}_{tg}+\boldsymbol{u}_{fb}.
\end{equation*}

One interpretation of this equation is that the \tg{} generates possibly sub-optimal actions based on parameters learned by the policy. To improve upon these sub-optimal actions, the policy learns correction terms that are added to the output or learns to suppress $u_{tg}$ if needed. 

A different, yet important, interpretation is that the policy optimization algorithm can use the \tg{} as a memory unit.
Because we do not place restrictions on the type of \tg{}, it makes sense to think about very simple choices of \tg{}s that still provide the policy with useful memory. For example, imagine choosing a leaky integrate-and-fire neuron~\cite{gerstner2002spiking} with a constant input as the \tg{} and letting the policy control the integration leak rate. In this case, the policy could use the \tg{} as a controllable clock signal. Because of this last interpretation, we only consider feed-forward neural networks for the policy in this work. All the memory is to be provided by the \tg{}. As we will demonstrate using both the synthetic control problem and robot locomotion, these benefits of the \tg{} allow us to efficiently learn architecturally simple policies (e.g. linear) that still generate complex and robust behavior.

\begin{figure}[h]
\centering
    \begin{minipage}{.5\textwidth}
      \centering
      \includegraphics[width=.8\linewidth]{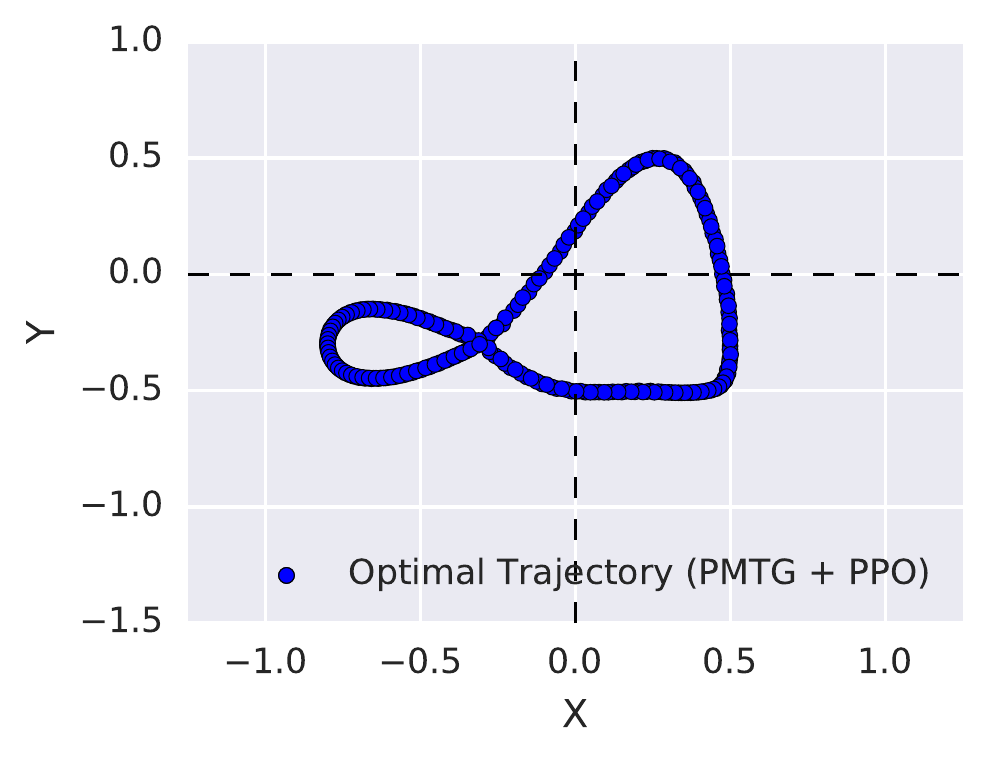}
      \subcaption{Optimal (learned) behavior.}
      \label{fig:pointmass-result}
    \end{minipage}%
    \begin{minipage}{.5\textwidth}
      \centering
      \includegraphics[width=.8\linewidth]{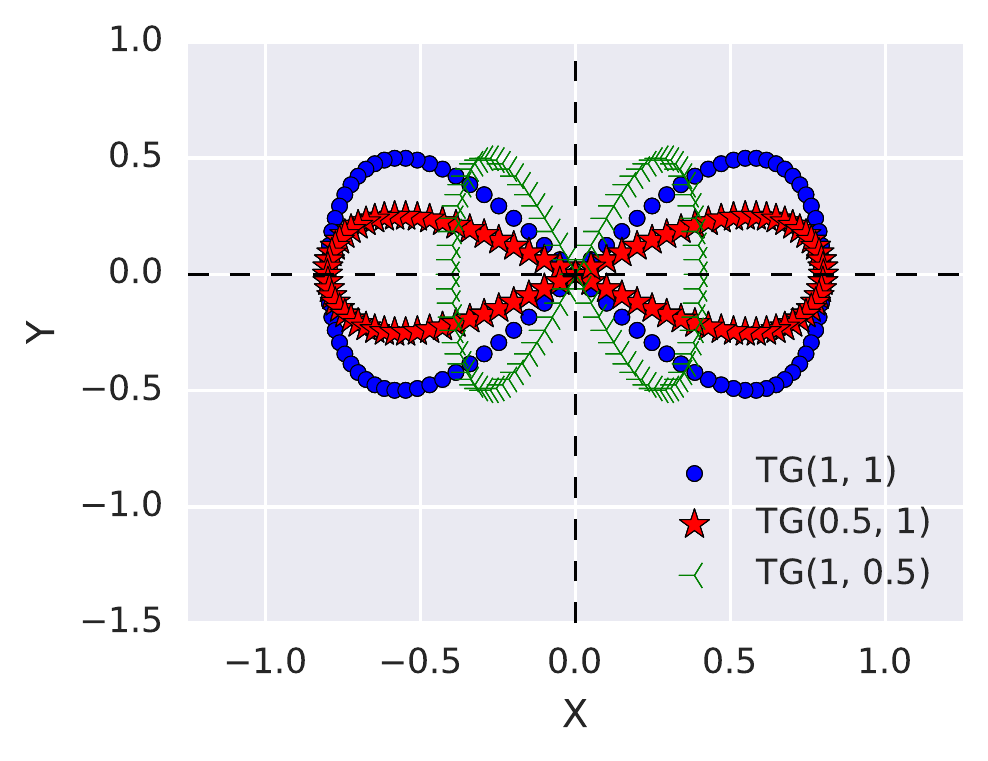}
      \subcaption{\tg{} with 3 different sets of parameters.}
      \label{fig:pointmass-TG}
    \end{minipage}
    \caption{2D Synthetic Control problem with a desired pattern and the \tg{} that generates figure-eights.}
\end{figure}

We now introduce a synthetic control problem to illustrate how \pmtg{} works. We consider a 2D environment of \SI{2}{\meter}~by~\SI{2}{\meter} in which a point is to be moved along a desired cyclic trajectory to maximize the returned reward (Fig.~\ref{fig:pointmass-result}.
The input to the environment (action space) is the desired next position $\boldsymbol{u}=\left[\begin{matrix}u_x\, u_y\end{matrix}\right]^T$.

\begin{figure}[th]
      \centering
      \includegraphics[width=0.5\columnwidth]{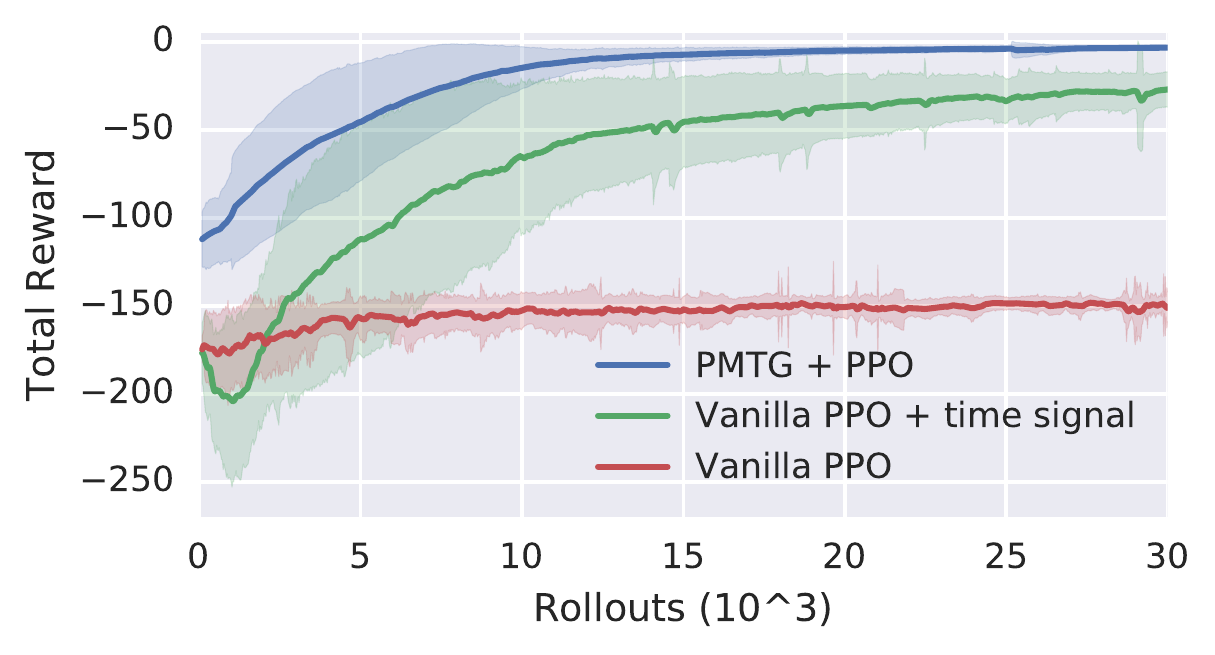}
      \caption{Learning curves for the 2D Synthetic Control Problem using \pmtg{} + PPO and Vanilla PPO. For Vanilla PPO, we simply remove the \tg{} from the setup ($\boldsymbol{u}=\boldsymbol{u}_{fb}$, $\boldsymbol{s}=\left[x,y\right]^T$.)}
      \label{fig:pointmass-training-react}
\end{figure}

As prior knowledge, under the assumption that we know that the trajectory will be a highly deformed and displaced version of figure-eight, we pick an eight curve as the trajectory generator and allow the policy to change the amplitudes along the $x$ and $y$ axes\footnote{To limit the complexity of this example, we do not use an external control signal nor allow the policy to control the offset or frequency of the trajectory generator.} (Fig.~\ref{fig:pointmass-TG}): 
\begin{equation*}
\boldsymbol{u}_{tg}(a_x, a_y)=
\left[\begin{matrix}
a_x \sin(2\pi t) \\
\frac{a_y}{2} \sin(2\pi t)\cos (2\pi t)
\end{matrix}
\right],
\end{equation*}
where $t$ represents the current timestep and is stored by \tg{} as its internal state.

For the policy, the observations are the current position along $x$ and $y$ coordinates and the state of the \tg{} (the current time step). The actions are the desired position $\boldsymbol{u}_{fb}$ and the parameters of the \tg{} $a_x, a_y$ (amplitudes used for the figure-eight). The reward is the negative Euclidian distance to a deformed figure-eight. We used the Proximal Policy Optimization (PPO) algorithm for learning with a fully connected neural networks with two layers and ReLU non-linearities~\cite{DBLP:journals/corr/SchulmanWDRK17}~\footnote{With hyperparameter search for up to 200 neurons per layer.}.

Using the architecture, \pmtg{} + PPO reaches almost optimal behavior with a reward close to zero (Fig.~\ref{fig:pointmass-training-react}). For comparison, training a pure reactive controller using Vanilla PPO fails to produce any good result. The failure to learn by the reactive controller can be explained by the nature of the task, partially observable state space, and lack of memory to distinguish different phases of the target figure. Since the reactive controller lacks time-awareness (or external memory), we also tested Vanilla PPO with a time signal as an additional observation. This combination performed better than Vanilla PPO, but still worse than \pmtg{}. In this example problem we showed a basic \tg{}, its combination with a feed-forward policy, and how \pmtg{} allows a feed-forward policy to learn a problem that is challenging for a pure reactive controller.

\section{Quadruped Locomotion}
\label{sec:quadruped}

\subsection{Controller Design}

\begin{wrapfigure}{r}{6cm}
      \centering
      \includegraphics[width=6cm]{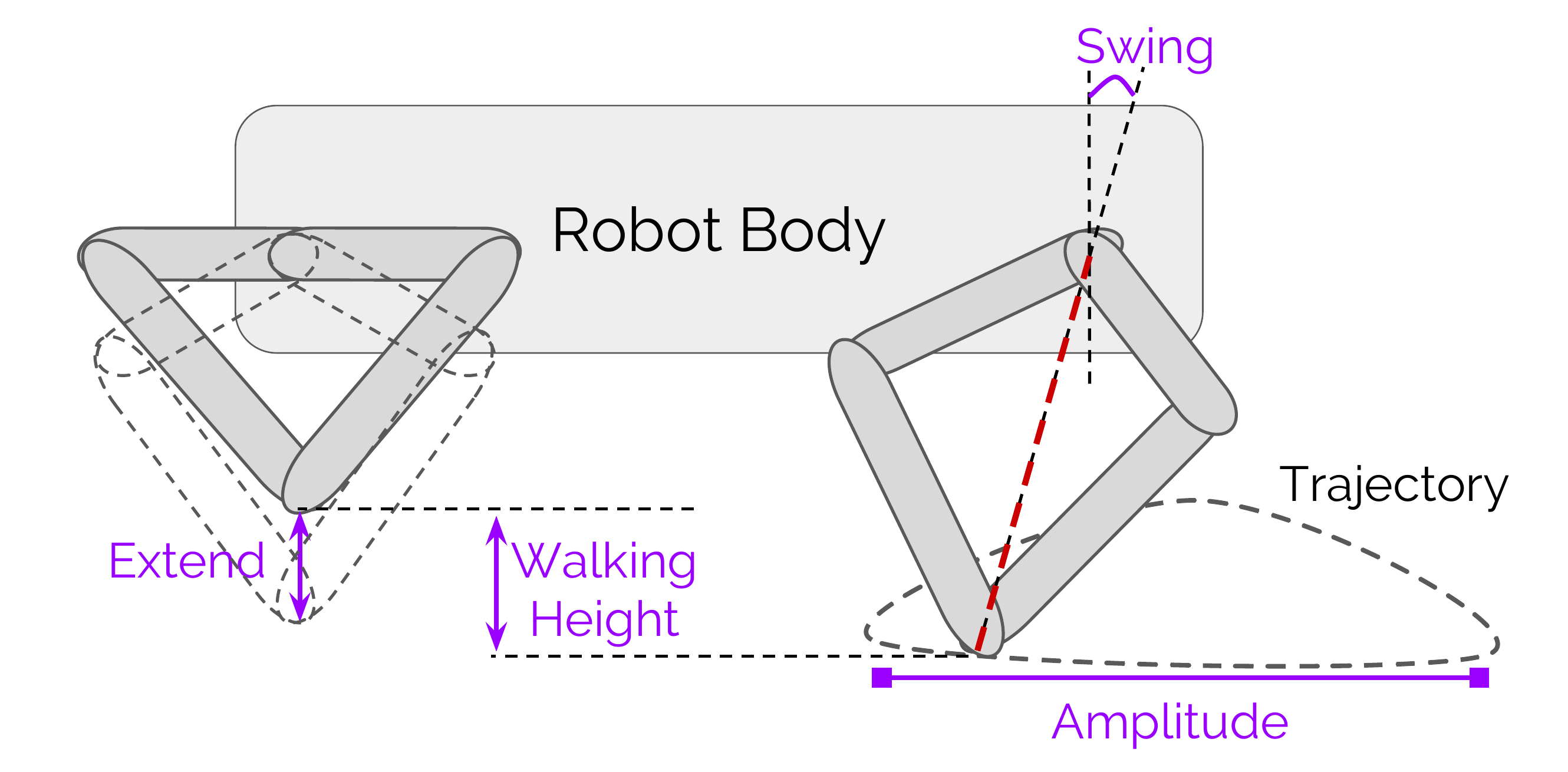}
      \caption{Illustration of robot leg trajectories generated by the \tg{}.} 
      \label{fig:robot-leg}
\end{wrapfigure}
Robot locomotion is a challenging problem for learning. Partial observations, noisy sensors combined with latency, and rich contacts all increase the difficulty of the task. Despite the challenges, the nature of locomotion makes it a good fit for \pmtg{}. \tg{} design can be based on the idea that legs follow a periodic motion with specific characteristics. A clear definition of the legs' trajectories is not needed. Instead, we can roughly define the family of trajectories using parameters such as stride length, leg clearance, walking height, and frequency. Fig.~\ref{fig:robot-leg} shows a sample trajectory of the leg and parameters based on this idea. The detailed definition of a \tg{} for locomotion can be found in Appendix.

\begin{figure}[th]
      \centering
      \includegraphics[width=0.7\columnwidth]{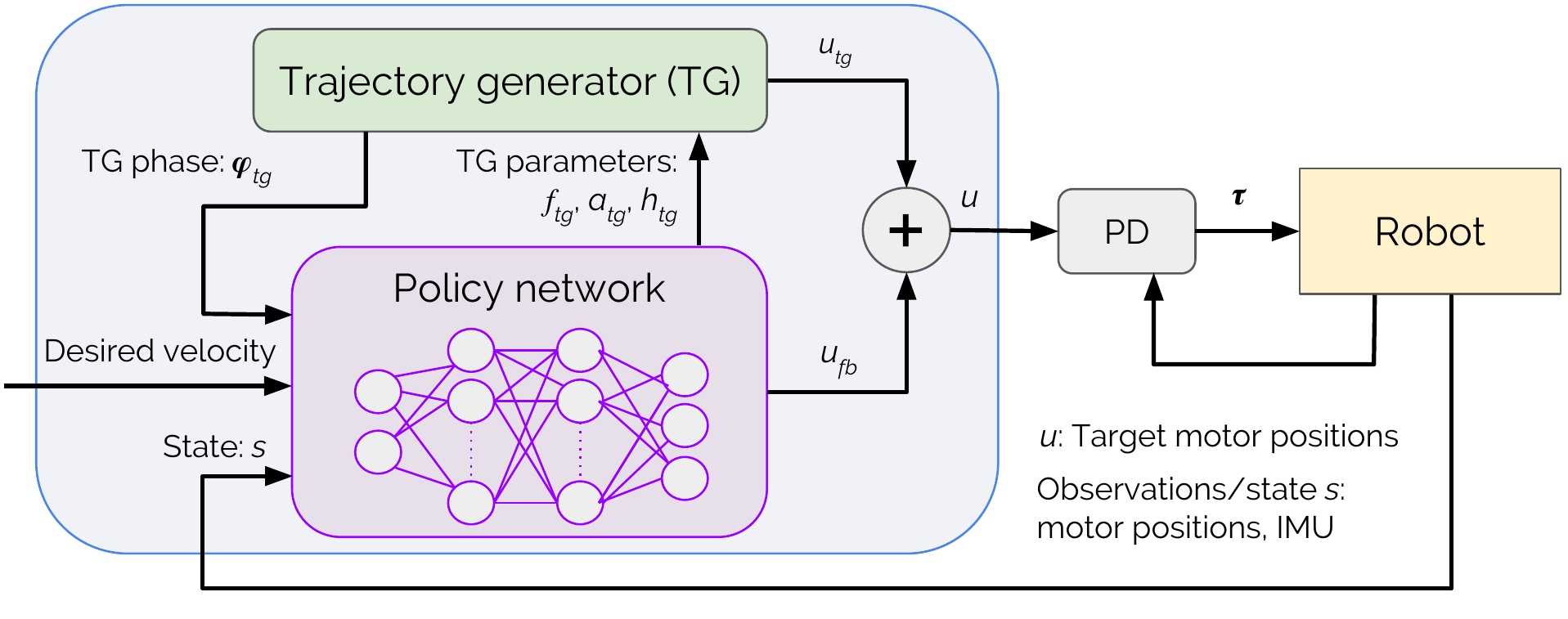}
      \caption{Adaptation of \pmtg{} to the quadruped locomotion problem.}
      \label{fig:architechture-minitaur}
\end{figure}

The detailed architecture adapted to quadruped locomotion is shown in Fig.~\ref{fig:architechture-minitaur}. At every timestep, the policy receives observations ($\boldsymbol{s}$), desired velocity ($v$, control input) and the phase ($\phi$) of the trajectory generator. It computes 3 parameters for the \tg{} (frequency $f$, amplitude $a$ and walking height $h$) and 8 actions for the legs of the robot ($\boldsymbol{u}_{fb}$) that will directly be added to the \tg{}'s calculated leg positions ($\boldsymbol{u}_{tg}$). The sum of these actions is used as desired motor positions, which are tracked by Proportional-Derivative controllers. Since the policy dictates the frequency at each time step, it dictates the step-size that will be added to \tg{}'s phase. This eventually allows the policy to warp time and use the \tg{} in a time-independent fashion. 

\begin{figure}[t]
\centering
    \begin{minipage}{.5\textwidth}
      \centering
      \includegraphics[width=.8\linewidth]{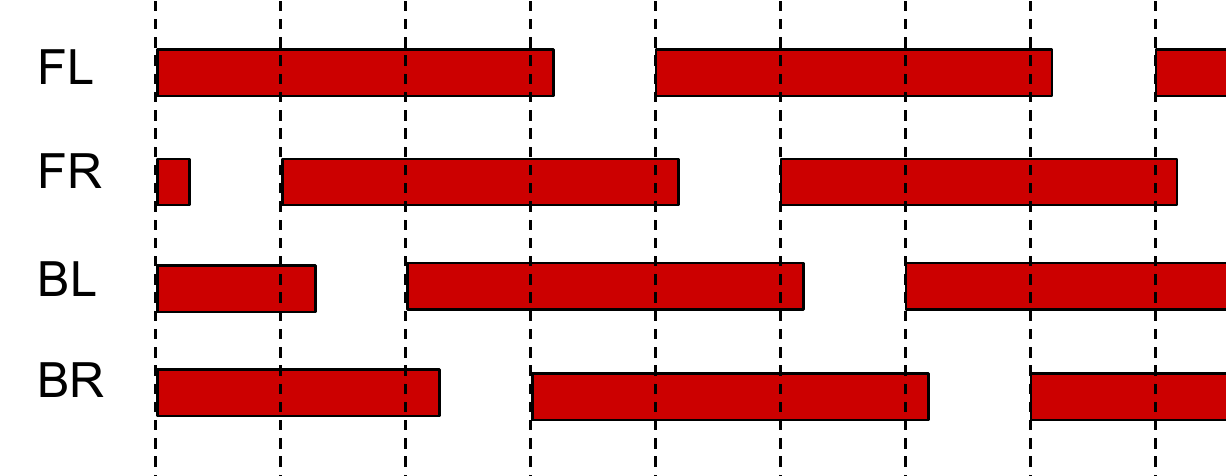}
      \subcaption{Leg phases of \tg{} for walking and running.}
      \label{fig:gait-walk}
    \end{minipage}%
    \begin{minipage}{.5\textwidth}
      \centering
      \includegraphics[width=.8\linewidth]{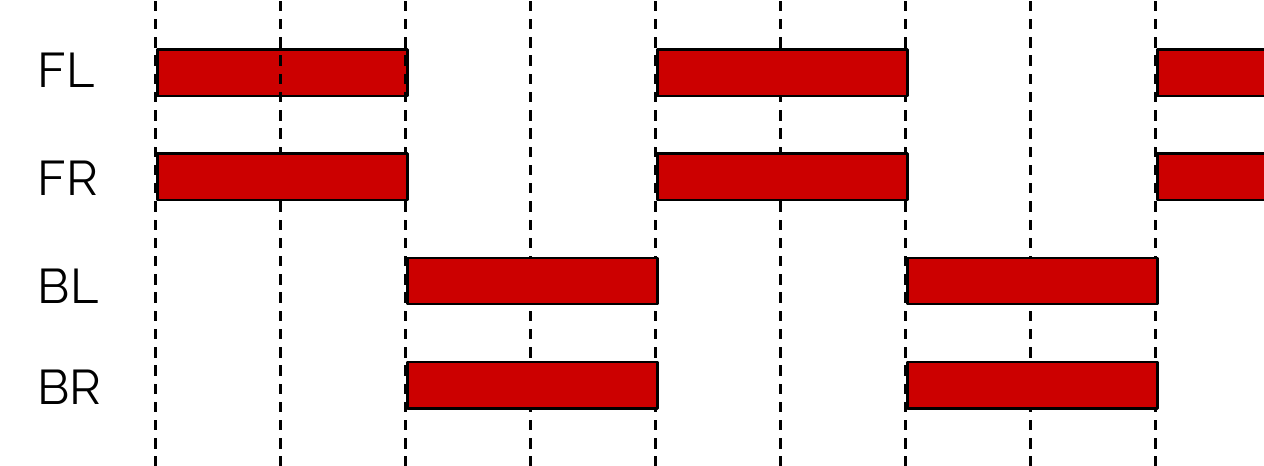}
      \subcaption{Leg phases of \tg{} for bounding.}
      \label{fig:gait-bounding}
    \end{minipage}
    \caption{Leg phases for 2 different gaits used for \tg{}. For each leg, the red bar indicates the duration of the stance (compared to swing represented by empty areas). FL: Front Left, FR: Front Right, BL: Back Left, BR: Back Right.}
    \label{fig:gaits}
\end{figure}

For locomotion, the design of \tg{} can be as simple as Fig.~\ref{fig:robot-leg} or can be composed of more complex open loop mechanisms. We use a stateful module that generates trajectories in an open loop fashion by using 3 parameters (walking height, amplitude, frequency). It is possible to use a \tg{} that is pre-optimized for the given task, or hand-tuned to roughly generate a desired gait. For walking and running gaits, we used a \tg{} that uses a gait shown in Fig.~\ref{fig:gait-walk} and pre-optimized as a standalone open-loop controller. Despite the pre-optimization, the \tg{} itself cannot provide stable forward locomotion since it lacks the feedback from the robot. In addition, for the bounding gait, we tested \pmtg{} with a simpler and a hand-tuned \tg{} that is not optimized. The only behavior provided by the \tg{} is swinging front and back legs in half period phase difference (Fig.~\ref{fig:gait-bounding}).

\subsection{Training}
We train the locomotion policy using PyBullet~\cite{coumans2018} to simulate the Minitaur robot from Ghost Robotics. As the training algorithm we use both Evolutionary Strategies (specifically ARS \cite{ars2018}) and Reinforcement Learning (specifically PPO \cite{DBLP:journals/corr/SchulmanWDRK17}). 
During training, we vary the desired forward velocity during each rollout. We start with \SI{0}{\meter\per\second}, gradually increase the desired speed to \SI{0.4}{\meter\per\second}, and keep it there for a while, then decrease it back to \SI{0}{\meter\per\second} by the end of the rollout. The exact speed profile is shown in Fig.~\ref{fig:minitaur-velocity}. During each rollout, We add random directional perturbation forces (up to 60N vertical, 10N horizontal) to the robot multiple times to favor more stable behaviors. Each rollout ends either at \SI{25}{\second} or when the robot falls. The reward function is calculated based on the difference between the desired speed and robot's speed as:
\begin{equation*}
    R = v_{\max} e^{-\frac{(v_R - v_T)^2}{2 {v_{\max}}^2}},
\end{equation*} 
where $v_{\max}$ is the maximum desired velocity for the task, $v_R - v_T$ are the robot's actual velocity and the target velocity at the current timestep. We selected this reward function because it provides the maximum reward when the robot is within the range (20\% of the top speed) of the desired speed and decreases to \SI{0}{\meter\per\second} if the difference is higher.

The observation includes the robot's roll and pitch and angular velocities along these two axes received (IMU sensor reading, 4 dimensions total). Overall the policy uses 7 input dimensions: 4 observation dimensions, the desired velocity as the control input, and the phase of the \tg{} represented by $sin(\phi), cos(\phi)$. The action space for the policy is 11 dimensional: 8 dimensions for the legs (swing and extension for each leg), and 3 parameters consumed by the \tg{} (frequency, amplitude, walking height).

For policy complexity, we evaluated both a two-layer fully connected neural network (up to 200 neurons per layer) as well as a simple linear policy (77 parameters). We trained the policies using 3 separate tasks: slow walking (up to \SI{0.4}{\meter\per\second}), fast walking (up to \SI{0.8}{\meter\per\second}) and bounding (up to \SI{0.4}{\meter\per\second}). For the bounding gait, we use a different \tg{} with phases shown in Fig.~\ref{fig:gait-bounding}. For walking gaits, the \tg{} alone does not provide forward motion, but the robot does not immediately fall. The open loop (\tg{} only) bounding gait fails immediately. 

\subsection{Results}
\begin{figure}[t]
\centering
    \begin{minipage}{.5\textwidth}
      \centering
      \includegraphics[width=1.0\linewidth]{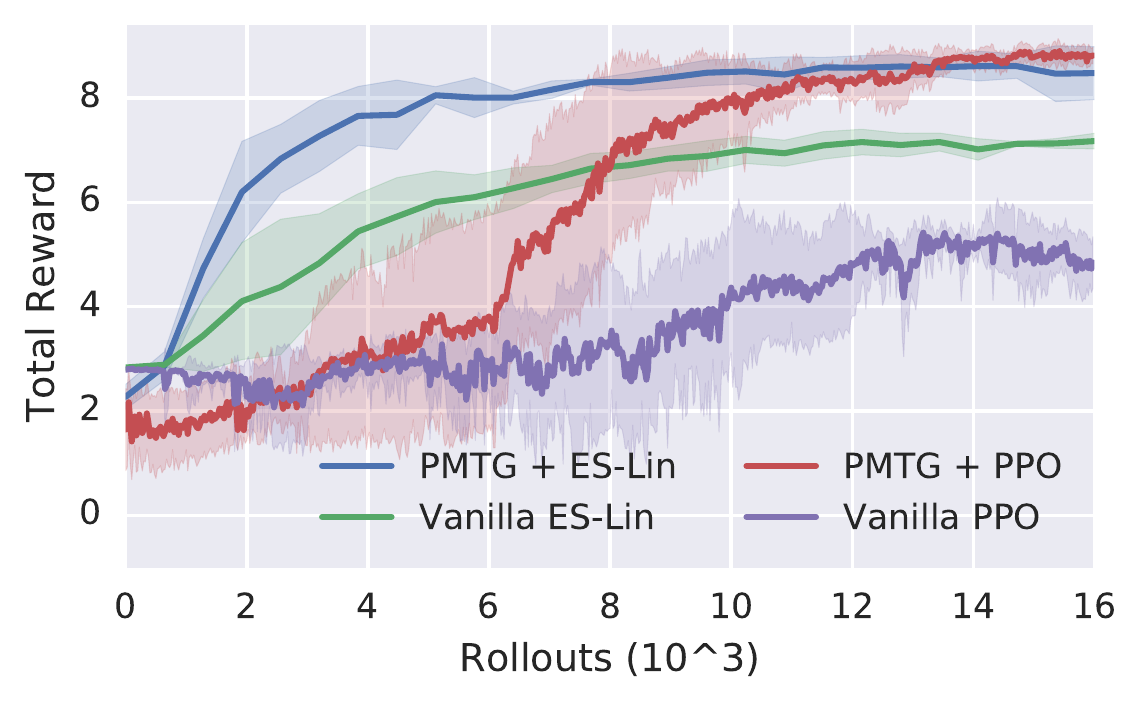}
      \subcaption{Learning with and without \pmtg{}.}
      \label{fig:minitaur-training}
    \end{minipage}%
    \begin{minipage}{.5\textwidth}
      \centering
      \includegraphics[width=1.0\linewidth]{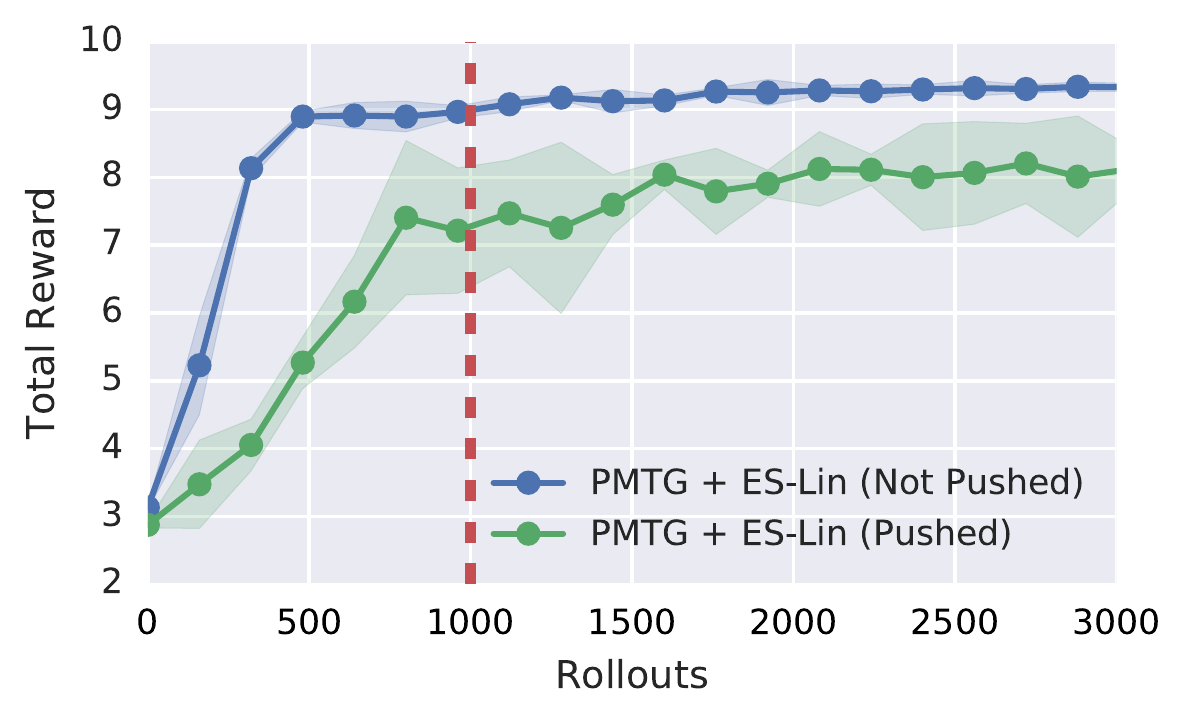}
      \subcaption{Fastest learning configurations (rollouts $<$ 1000).}
      \label{fig:minitaur-training-fast}
    \end{minipage}
    \caption{Training curves for quadruped walking with speed tracking using \pmtg{}. Left: Learning curves of \pmtg{} vs. PPO and ES. Right: Examples of experiments  that learned the desired gait in fewer than $1000$ rollouts (ES with Linear Policy using 8 directions per iteration).}
\end{figure}

Our architecture makes learning the complex locomotion task easier. When we use \pmtg{}, both algorithms successfully learn controllable locomotion (Fig.~\ref{fig:minitaur-training}). Both the linear controller and the two-layer feed-forward neural network achieve the desired behavior. The curves for Vanilla ES-Lin and Vanilla PPO show the results for a reactive controller instead of \pmtg{} (we simply remove the \tg{}, $\boldsymbol{u}=\boldsymbol{u}_{fb}$). Without \pmtg{} both algorithms fail to achieve the optimal reward levels. Lower rewards show that the controller learns the walking behavior but cannot fully keep up with the changing target speed \footnote{The literature contains successful learning of reactive controllers on locomotion tasks with less complexity, richer state space and different reward functions~\cite{tan2018}.}.

By combining \pmtg{} with ARS and a linear policy, we achieved high data efficiency for learning locomotion. We also note that the linear policy has relatively few  parameters~(77). As an added benefit, we observed that learning with \pmtg{} combined with a linear policy required fewer rollouts for the given locomotion task. 
Fig.~\ref{fig:minitaur-training-fast} shows learning curves for the hyperparameters with the fastest learning speed (ES with 8 directions per iteration). We observe that it is possible to learn good policies with ES in fewer than 1000 rollouts. This is possible because we were able to embed prior knowledge into the \tg{} and because the architecture reduces the complexity of the policy learning problem.
The number of rollouts is low relative to the complexity of the locomotion task. This opens a research direction to using \pmtg{} for on-robot learning, which we are planning as a future work. 

\begin{figure}[t]
\centering
    \begin{minipage}{.99\textwidth}
      \centering
      \includegraphics[width=.8\linewidth]{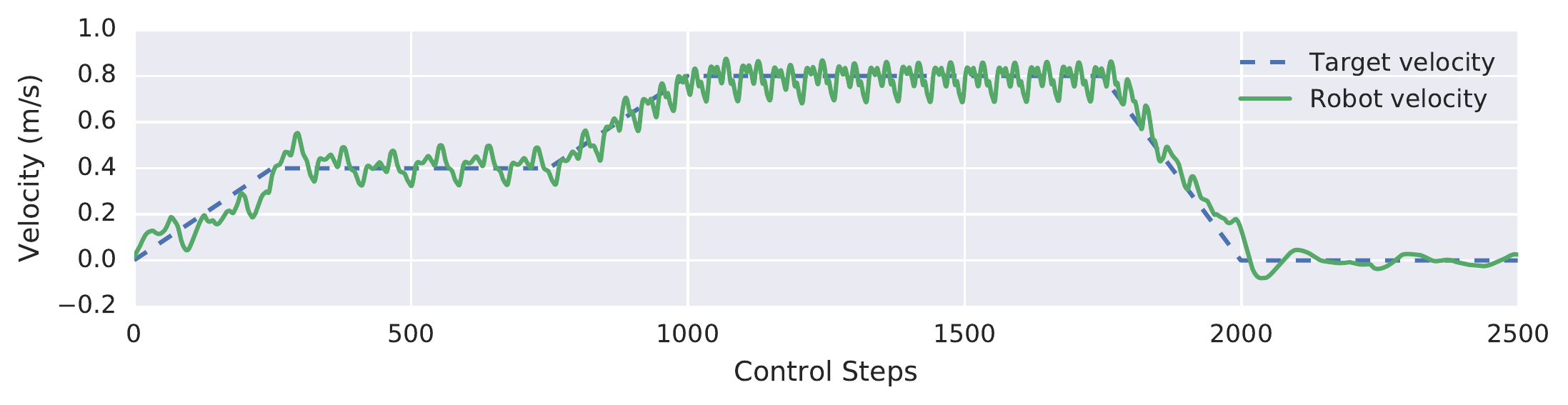}
      \subcaption{Speed profile.}
      \label{fig:minitaur-velocity}
    \end{minipage}%
    \\
    \begin{minipage}{.99\textwidth}
      \centering
      \includegraphics[width=.8\linewidth]{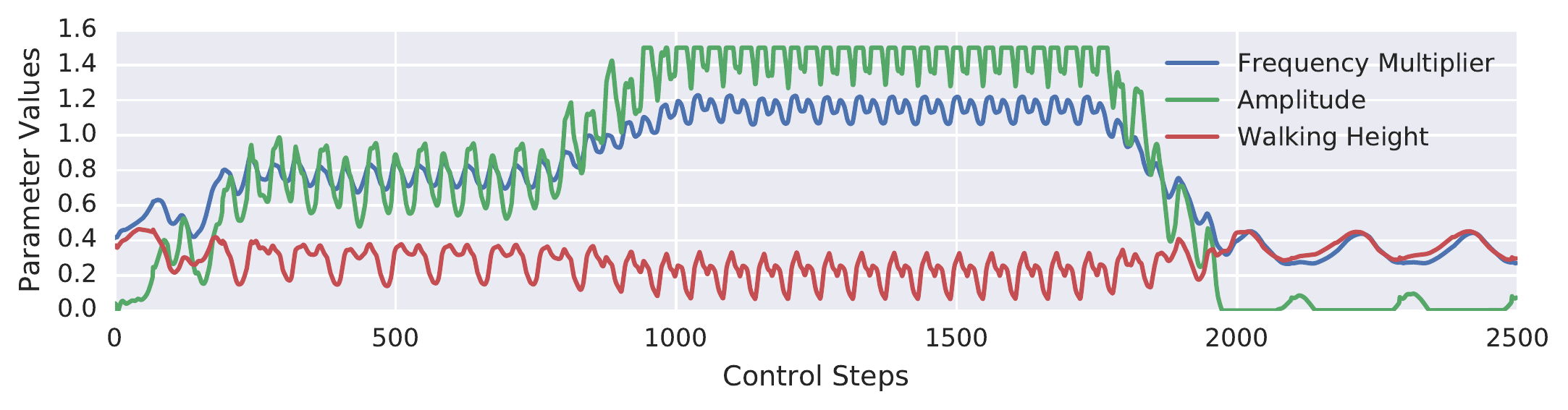}
      \subcaption{\tg{} parameters selected by the policy.}
      \label{fig:minitaur-tgparams}
    \end{minipage}
    \\
    \begin{minipage}{.99\textwidth}
      \centering
      \includegraphics[width=.8\linewidth]{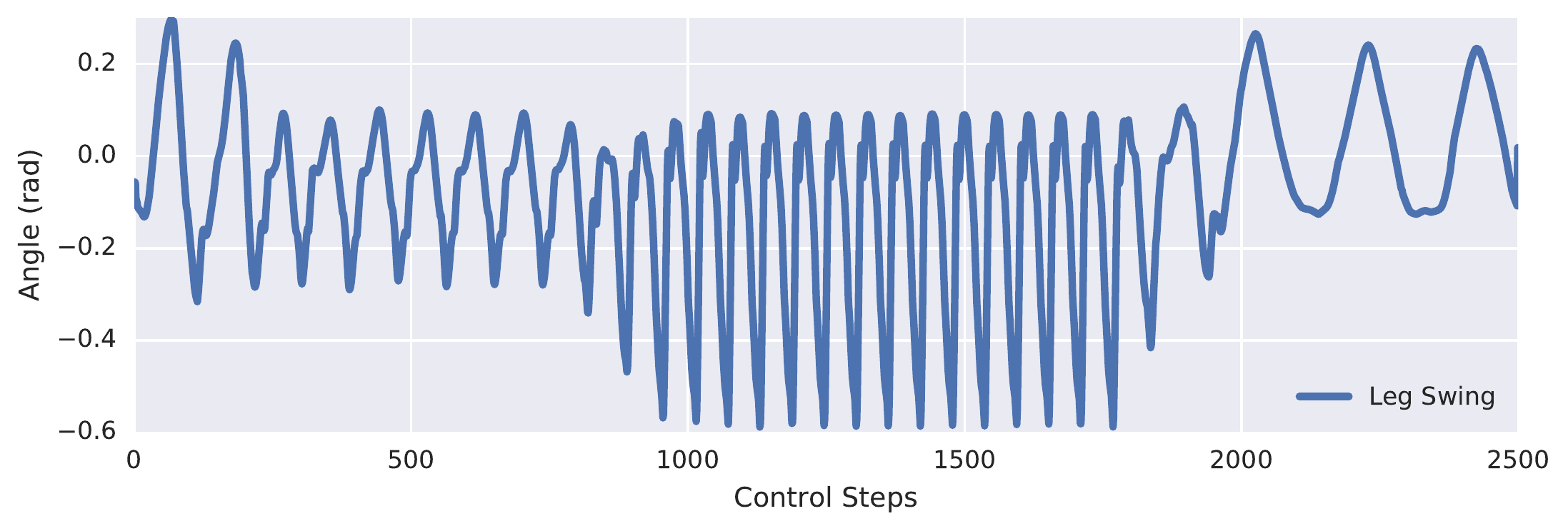}
      \subcaption{Swing angle of one of the legs.}
      \label{fig:minitaur-legswing}
    \end{minipage}
    \caption{Characteristics of a converged policy during running. The robot follows the change in desired speed while changing the parameters during the course. The plot of leg swing angle shows major changes in the emerged pattern at different speeds.}
\end{figure}

Next, we look at the characteristics of a sample converged controller using \pmtg{} and ES with a linear policy. We focus on running instead of walking because it shows considerable changes in \tg{} parameters and gait during a single rollout. Fig.~\ref{fig:minitaur-velocity} shows a single run after training: The robot does not have any problems tracking the desired speed. Fig.~\ref{fig:minitaur-tgparams} shows that the policy significantly modulates both the amplitude (which commands stride length) and the frequency of the gait depending on the desired speed. These parameters affect the output coming from the \tg{}, but they do not necessarily show the eventual leg movement since the policy can add corrections. Fig.~\ref{fig:minitaur-legswing} shows the swing angle of one the legs during the same rollout. The motion of the leg is periodic, but the shape of the signal changes significantly depending on the speed.

\subsection{Robot Experiments}
The reality gap between simulation and real environments is a major challenge for learning in robotics. In many scenarios, learning in simulation can easily converge to unrealistic behaviors or exploit the corner cases of a simulator. In \pmtg{}, we provide a class of initial trajectories (\tg{}) that the policy can build upon. The converged policies usually follow the characteristics of the \tg{}, avoiding unrealistic behaviors. Additionally, we use randomization by applying random directional virtual forces to the robot during training to avoid overfitting to the simulation model. 

We deployed a number of the learned controllers to the robot to see how our results transfer to the real world. We define success if the robot successfully moves forward  at various speeds and does not fall during the rollout. A short summary of these results is shown in the supplementary video. For slower walking, all the policies successfully worked on the robot. The emerged behaviors are similar to the simulation. The robot slowly increases its speed and walks at different desired speeds and slows down to stop without any observable problems. 

The policies trained for walking at faster speeds (up to \SI{0.8}{\meter\per\second}) mostly completed the rollouts successfully, but the legs were occasionally slipping at higher speeds. Although slippage affected the overall behavior for certain policies (i.e. distance run, direction) the robot recovered from falling and continued walking in most trials. 

When we used the \tg{} with a bounding gait, we observed different emerged gaits for different learning algorithms and hyperparameters. The policies trained with PPO were the most stable, jumping forward by modulating the parameters for walking height. The policy significantly overrode the gait using its correction ability. The resulting behavior shows forward movement of the robot using jumps at different speeds. We also include these different behaviors in the supplementary video.

\begin{figure}[th]
      \centering
      \includegraphics[width=0.99\columnwidth]{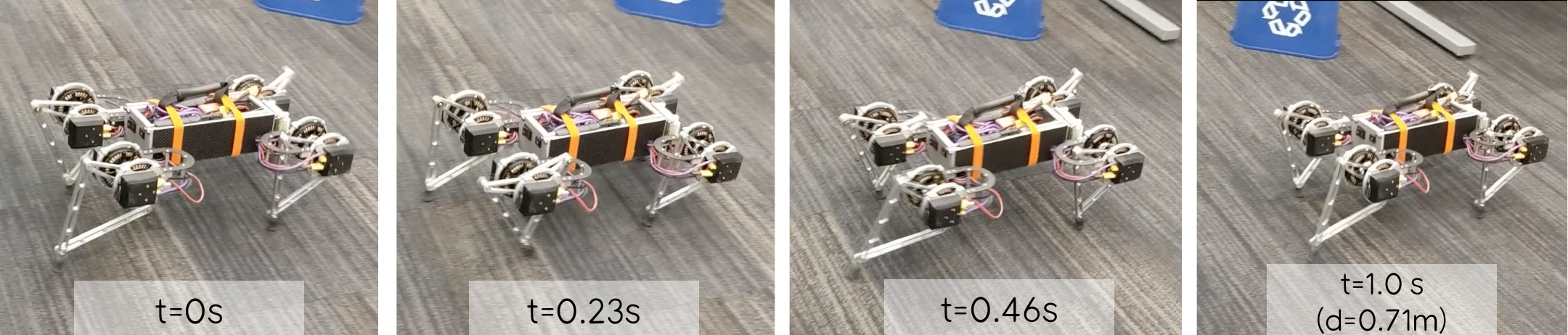}
      \caption{Minitaur robot walking using the learned controller.}
      \label{fig:minitaur_walking_2}
\end{figure}

\begin{figure}[th]
      \centering
      \includegraphics[width=0.99\columnwidth]{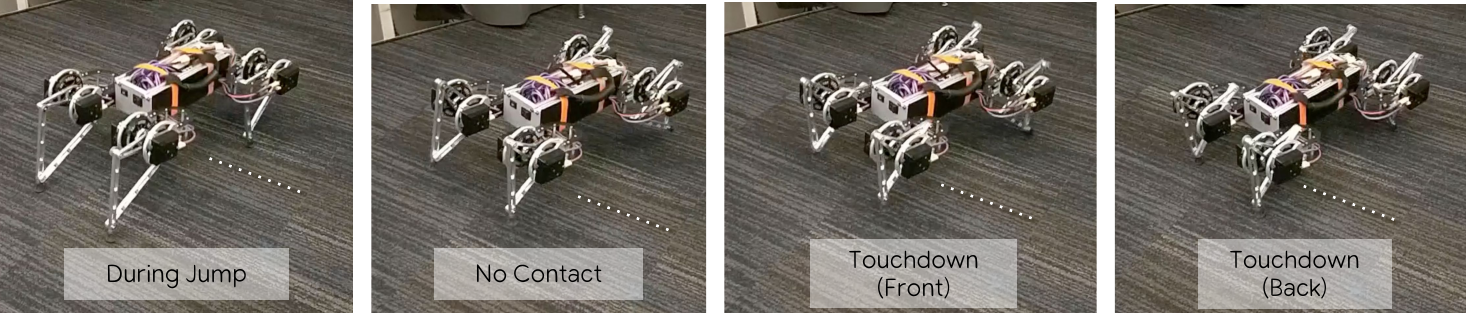}
      \caption{Minitaur robot trained with the \tg{} for bounding.}
      \label{fig:minitaur_walking_3}
\end{figure}


\section{Conclusion}
\label{sec:conclusion}
We introduced a new control architecture (\pmtg{}) that represents prior knowledge as a parameterized Trajectory Generator (\tg{}) and combines it with a learned policy. Unique to our approach, the policy modulates the trajectory generator at every time step based on observations coming from the environment and \tg{}s internal state. This allows the policy to generate many behaviors based on this prior knowledge, while also using \tg{} as a memory unit. This combination enables simple reactive policies (e.g. linear) to learn complex behaviors. 

To validate our technique, we used \pmtg{} to tackle quadruped robot locomotion. We show the generality of \pmtg{} by training the architecture using both ES and RL algorithms on two different gaits. We used relatively simple policies considering the complexity of the locomotion task, and had success with a linear policy. The policy uses only IMU readings -- a low dimensional set of observations for a robot locomotion task -- and takes the desired velocity as an external control input. We showed successful transfer of these policies from simulation to the Minitaur robot.

We plan to use our current approach as a starting point for this class of architectures with simple learned policies. 
In this work we relied on ad hoc trajectory generators that were chosen based on our intuition about a given problem. In future work, we are interested in getting a deeper understanding of which types of trajectory generators work best for a specific domain, possibly extracting trajectory generators from demonstrations. Finally, we are interested in theoretical foundations for \pmtg{} and how it relates to existing models, specifically recurrent ones.



\clearpage
\acknowledgments{The authors would like to thank the members of the Google Brain Robotics group for their support and help.}


\bibliography{main}  

\clearpage
\section*{Appendix}
The phase of the trajectory generator (between $0$ and $2\pi$) is defined as:
\begin{equation}
    \phi_t = \phi_{t-1} + 2 \pi f_{tg} \Delta t \textrm{   mod } 2 \pi,     
\end{equation}
where $f_{tg}$ defines the frequency of the trajectory generator. In \pmtg{} architecture, $f_{tg}$ is selected by the policy at each time step as an action.

In this work, we use the following trajectory generator for the legs:
\begin{equation}
\boldsymbol{u}_{tg} =
\left[\begin{matrix}
S(t) \\
E(t) 
\end{matrix}
\right]
=
\left[\begin{matrix}
C_s + \alpha_{tg} cos(t') \\
h_{tg} + A_e sin(t') + \theta cos(t') 
\end{matrix}
\right].
\label{eq:openloop}
\end{equation}
\begin{itemize}
    \item $S(t)$, and $E(t)$ are respectively the swing and extension of the leg as shown in Fig.~\ref{fig:robot-leg}.
    \item $C_s$ defines the center for the swing DOF and extension DOF (in radians).
    \item $h_{tg}$ defines the center for the extension DOF.
    Extension is represented in terms of rotation of the two motors in the opposite direction, hence the unit is also radians. Since all legs share the same $h_{tg}$, it corresponds to the walking height of the robot. 
    \item $\alpha_{tg}$ defines the amplitude of the swing signal (in radians). This corresponds to the size of a stride during locomotion.
    \item $A_e$ defines the amplitude of the extension during swing phase. This corresponds to the ground clearance of the feet during the swing phase.
    \item $\theta$ defines the extension difference between when the leg is at the end of the swing and when the leg is at end of the stance. This is mostly useful for climbing up or down.
\end{itemize}

We compute $t'$ based on the swing and stance phases:
\begin{equation}
    t' = \begin{cases}
         \frac{\phi_{\textrm{leg}}}{2(1 - \beta)} & \textrm{if }0 < \phi_{\textrm{leg}} < 2 \pi \beta; \\
        2\pi - \frac{(2\pi - \phi_{\textrm{leg}}) }{2\beta} & \textrm{otherwise,}
    \end{cases}
\end{equation}
where $\beta$ defines the proportion of the duration of the swing phase to the stance phase.

For each leg, the phase is calculated separately as
\begin{equation}
    \phi_{\textrm{leg}} = \phi_t + \Delta\phi_{\textrm{leg}} \textrm{ mod } 2 \pi, 
\end{equation}
where $\Delta\phi_{\textrm{leg}}$ represents the phase difference of this leg compared to the first (left front) leg. This is defined by the selected gait (i.e. walking vs bounding).
\end{document}